\documentclass[]{article}
\usepackage{bagrow}
\usepackage{amsmath}
\usepackage{amssymb}
\usepackage{amsthm}
\usepackage{graphicx}
\usepackage[]{newtxtext}
\usepackage[bigdelims]{newtxmath}
\usepackage{booktabs}
\usepackage[usenames,dvipsnames,svgnames,table]{xcolor}
\usepackage{xcolor}
\usepackage[normalem]{ulem}

\usepackage[final]{pdfpages}

\def\suppmat{Supporting Material}
\def\SM{SM}
\def\rrep{r}
\def\rpre{\hat{r}}

\newcommand\avg[1]{\ensuremath{\mathrm{E}\left[ #1 \right]}}

\newcommand{\thecorr}{0.8201}

\begin{document}

\author[1,2,*]{James P.~Bagrow}
\author[3,2]{Daniel Berenberg}
\author[3,2]{Joshua Bongard}
\affil[1]{Department of Mathematics \& Statistics, University of Vermont, Burlington, VT, United States}
\affil[2]{Vermont Complex Systems Center, University of Vermont, Burlington, VT, United States}
\affil[3]{Department of Computer Science, University of Vermont, Burlington, VT, United States}
\affil[*]{\corrauthinfo{james.bagrow@uvm.edu}{bagrow.com}}

\title{Neural language representations predict outcomes of scientific research}

\date{May 17, 2018}


\maketitle

\begin{abstract}
Many research fields codify their findings in standard formats, often by reporting correlations between quantities of interest.
But the space of all testable correlates is far larger than scientific resources can currently address, so
the ability to accurately predict correlations would be useful to plan research and allocate resources.
Using a dataset of approximately 170,000 correlational findings extracted from leading social science journals,
we show that a trained neural network can accurately predict the reported correlations using only the text descriptions of the correlates. 
Accurate predictive models such as these can guide scientists towards promising untested correlates, better quantify the information gained from new findings, and has implications for moving artificial intelligence systems from predicting structures to predicting relationships in the real world.
\end{abstract}



%

\section{Introduction}

One of the most important applications of machine learning is its ability to replace data that are difficult, expensive or dangerous to collect with predictions generated using more amenable, economical or ethical data.
Examples of this include replacing unavailable socioeconomic indicators for regions that are difficult to survey with predictions made from satellite imagery~\cite{jean2016combining},
predicting pneumonia mortality and hospital readmission~\cite{cooper1997evaluation,caruana2015intelligible},
inferring cancer risk from medical imagery or histology results~\cite{cirecsan2013mitosis,cruz2014automatic,xu2016deep}.
Predictions learned from data can also guide the scientific discovery process.
Some examples include predicting novel chemical reactions from previously collected experimental data~\cite{raccuglia2016machine}, classifying quasar candidates and estimating stellar parameters from photometric data~\cite{richards2004efficient,fiorentin2007estimation}, or
particle discovery from high-energy physics experiments~\cite{baldi2014searching}.

Correlational findings underpin significant portions of the scientific enterprise, and inform statistical methods, study design and meta-analyses~\cite{bosco2015correlational}. 
Reported correlations are often the primary outcome of an experimental or empirical study and capture a field's ability to measure and explain the relationships they study.
Findings are often presented as numeric tables along with \emph{metadata}, the written descriptions of the correlates.
How much information is present in these written descriptions?
Can we quantify the relationships captured by the scientific record by predicting the correlation from the correlates only?
The ability to make actionable predictions would be invaluable to 
deriving new information from metadata contained within the scientific record,
identifying correlations unreported in the literature, 
and thus perhaps providing a guide for future studies.

\begin{figure}[t]
\centering
\includegraphics[width=\textwidth]{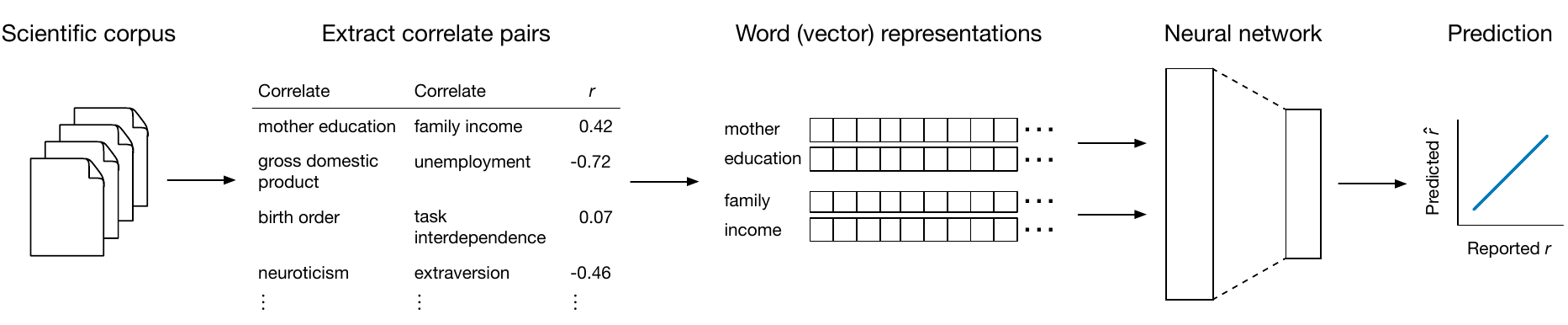}
\caption{Predicting correlational findings from written descriptions only.
Using a dataset of correlate pairs extracted from published manuscripts~\cite{bosco2015correlational,bosco2015cloud} and pretrained vector representations encompassing semantic knowledge of individual words~\cite{speer2017conceptnet}, we train a recurrent neural network to predict correlations as a function of correlate text.
\label{fig:overview}
}
\end{figure}

Our research process aiming to address these questions is illustrated in Fig.~\ref{fig:overview}.
We study a longitudinal dataset of approximately 170,000 correlate pairs, extracted from 30 years of research findings published in leading social science journals~\cite{bosco2015correlational,bosco2015cloud}.
These data provide the text descriptions of each correlate and the reported correlation.
We then train a recurrent neural network to predict a correlation $\rpre{}$ given only the ordered text sequences describing two correlates. 
While predictions can be made directly from words (tokens) taken from the correlates, the current practice in natural language processing is to embed tokens into a learned high-dimensional vector space~\cite{collobert2008unified,mikolov2013distributed,pennington2014glove}.
These vector representations of the tokens allow increased flexibility, for example by handling typos, and capture language syntax and semantics in a computationally useful manner~\cite{bengio2003neural}.
The recurrent neural network then learns relationships between sequences of these vectors that relate to the reported correlation.
Full details on text processing, network architecture, hyperparameters, and training are given in the \suppmat{}.

Our neural network uses
vector representations of words that are computed in an unsupervised manner from large-scale text corpora~\cite{mikolov2013distributed}.
As such, these representations will reflect any biases latent to those text, such as racial or gender bias~
\cite{bolukbasi2016man,Caliskan183}. 
This is especially problematic as most training corpora are gathered from uncurated text sources, typically web crawls known to be biased in a number of dimensions~\cite{Caliskan183}.
Given these biases, it is not clear how well such representations can support predictions of objective research findings.
Therefore, with these issues in mind, we utilize a pretrained vector representation called ConceptNet Numberbatch.
Numberbatch is an ensemble of multiple, state-of-the-art representations that has been further enhanced in two important ways: one, the representations are endowed with information from the ConceptNet knowledge graph, a long-running project to codify objective relationships between entities~\cite{speer2012representing}; and two, Numberbatch is explicitly trained to perform well at natural language tasks while also minimizing a number of bias indicator scores, such as the implicit association test. 
Numberbatch is currently the most competitive and least biased
vector representation available to researchers, making it the most suitable choice for our task~\cite{speer2017conceptnet}.

\section{Results}

\begin{figure}[t]
\centering
\includegraphics[width=0.45\textwidth]{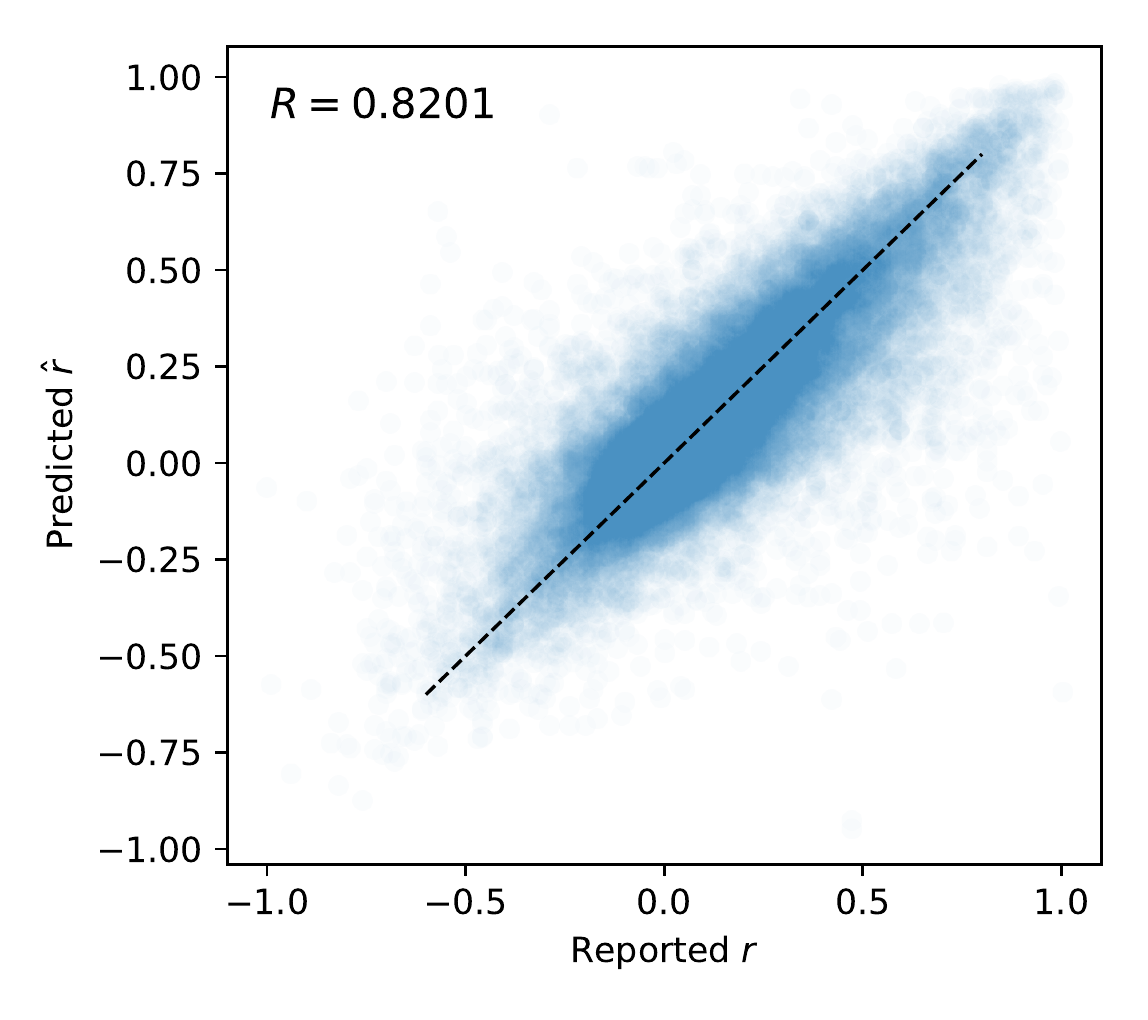}
\caption{Comparison of reported and predicted correlations for correlate pairs reserved for testing. 
Predicted values $\hat{r}$ show a strong correlation with reported values $r$: $R(\rrep{}, \rpre{}) \approx \thecorr$. The dashed line indicates $\rrep{} = \rpre{}$ and provides a guide for the eye.
\label{fig:scatterplot}
}
\end{figure}

To evaluate our ability to predict correlational findings, we trained the neural network on a random 80\% of the reported findings and reserved 20\% for testing purposes. 
After training, the neural network was asked to predict
the held-out correlations. 
As shown in Fig.~\ref{fig:scatterplot}, the model achieves accurate predictions, giving a correlation $R \approx \thecorr$ between reported $\rrep$ and predicted $\rpre$.

Given the accuracy demonstrated in Fig.~\ref{fig:scatterplot},
it is important to ask if the predictive model is learning meaningful relationships between correlates or if it is simply memorizing features implicit to the training corpus.
Memorization harms the ability of an algorithm to make accurate predictions.
While held-out or test data is the gold standard for evaluating predictive performance, and we used test data in Fig.~\ref{fig:scatterplot}, it is worthwhile to examine this issue further.
To do so, we introduce a comparative baseline predictive model: for any pair of correlates $c_i$ and $c_j$ we simply predict the mean correlation reported for any correlate pairs in the corpus that contain either $c_i$ or $c_j$. 
If the neural network is only memorizing the corpus, we expect this mean-value baseline to achieve comparable predictive performance.
Instead, the baseline performs significantly worse, with $R \approx 0.54$ (see \SM{}). 
This indicates that the neural network is learning meaningful representations and possesses predictive performance beyond this basic mean-value model.





The ability to make accurate predictions of correlational findings has many potential applications. Here we discuss two: (i) infilling a partial correlation table to interrelate a large set of correlates and (ii) discovering untested correlate pairs that are worth further investigation.

\begin{figure}
\centering
\includegraphics[width=\textwidth,trim=0 -7 0 0]{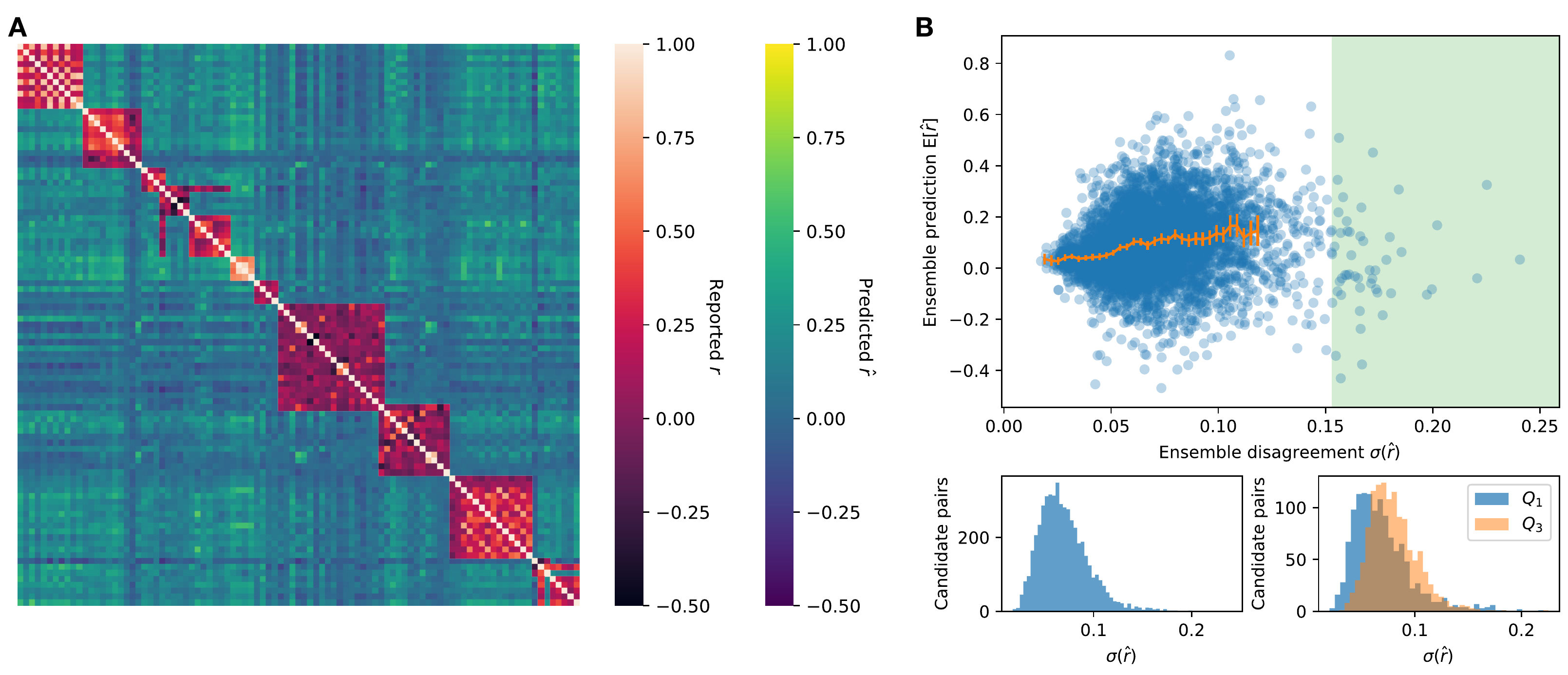}
\caption{Applications of correlation prediction.
\lett{A}
Infilling a multi-paper correlation table.
Reported correlations for 10 papers are shown along the diagonal in one color scale, predictions for
unreported correlations in another. Here 88.2\% of the table has been infilled by the predictive model.
\lett{B}
Finding candidate correlate pairs worth further investigation.
Candidate pairs with disagreement across an ensemble of predictive models are most worth testing as they will lowering that disagreement will yield the most information.
The highlighted region denotes the 1\% most uncertain candidate pairs.
We find higher disagreement for candidate pairs with more positive predicted correlation.
\label{fig:applications}
}
\end{figure}

Figure \ref{fig:applications}A shows a correlation table connecting ten papers published in 2010. The blocks along the diagonal of the table denote the reported findings of the individual papers; note that the third and tenth paper have some gaps as they do not report all correlations.
The table entries outside the blocks of reported findings and within the gaps of incomplete papers are those infilled by our predictive model.
These entries constitute 88.2\% of the table.
From this infilled table, we see that the predictive model is able to find meaningful connections across papers, in particular the first three papers and the last two papers show interesting correlations. 
These infilled relationships cannot be explained by the mean-value baseline model only (see \SM{}).
Researchers interested in topics bridging existing research can now receive preliminary guidance towards interesting convergence points.

Second, we ask if the model can find untested correlate pairs that are worth devoting resources into testing. 
The ability to guide research ahead of time is valuable as the space of possible relationships is very large.
Indeed, most correlate pairs in our corpus are untested:
of the 21,736 unique correlates recorded in the corpus (postprocessing), only 149,374 unique pairs have been tested, leaving 99.94\% untested.
Of course, many of these untested pairs will not be worthwhile or suitable to investigate, or have been tested in publications outside our corpus, yet there remain many correlate pairs to be studied and any guidance can help maximize limited research time.

Correlate pairs most worth further research are those that will provide the most novel information.
A classic approach to measuring information gain is from the prediction disagreement across an ensemble of predictive models: observations where the members of the ensemble give mostly the same prediction provide less novel information
than observations where members give diverse predictions. The latter case highlights valuable observations to learn from because the ensemble is more uncertain and gaining access to the correlation the ensemble is most uncertain about will yield the most novel information.
Seeking labels for training data via ensemble disagreement is an active learning technique known as \emph{Query by Committee}~\cite{seung1992query,krogh1995neural}.

We apply ensemble disagreement to highlight untested correlate pairs that may yield the most information when tested.
We trained an ensemble of $N = 50$ neural networks (see \SM{} for details) and then performed a simple random search: correlates were paired at random into $n = 5000$ candidate pairs and sent to each member of the network ensemble.
For each candidate correlate pair we measured the ensemble prediction $\avg{\rpre{}}$ and the ensemble disagreement $\sigma\left(\rpre{}\right)$, 
where $\avg{\cdot}$ and $\sigma(\cdot)$ denote the sample mean and sample standard deviation, respectively, taken over the predictions of the $N$ predictive models.

Figure~\ref{fig:applications}B shows the results of this search\sout{, comparing the ensemble disagreement to the ensemble prediction}. 
Correlate pairs with low {\color{red} ensemble} disagreement are well understood by the model ensemble, while those with high disagreement are not.
Query-by-committee argues that those correlate pairs that have high disagreement are most worth investigation, even if the predicted correlation is low, as those are most likely to  provide novel information gain if tested.
We highlight the 1\% most uncertain candidate pairs in Fig.~\ref{fig:applications}B.
Examining these candidate correlate pairs (see \SM{}), several interesting pairs appear (paraphrased here):
 `Gross domestic product' and `job search behavior (unemployed group)' had a predicted correlation of $\rpre{} = \avg{\rpre{}} \pm 1.96 \sigma\left(\rpre{}\right)/\sqrt{N}= -0.37 \pm 0.046$. 
Two more pairs with strong predicted correlations were `Manipulation check' and `Hardworking' 
($\rpre{} = 0.45 \pm 0.048$) and
`Overall handshake' and `Feels superior' ($\rpre{} = 0.34 \pm 0.043$).
These pairs are quite sensible: manipulation checks~\cite{oppenheimer2009instructional} are a common approach to filter out study participants who are not performing a task well, and handshakes and body language are often associated with feelings of superiority.
Another pair, `Standard Occupational Classification' and `Perceived interactional justice' ($\rpre{} = -0.43 \pm 0.044 $) may highlight some research directions relating how people can interact positively with co-workers.
Lastly, the pair `Organizational embeddedness' and `Novice completion time' ($\rpre{} = -0.04 \pm 0.061 $) shows a marginal correlation effect but the high degree of disagreement may demonstrate the need for more research on how organizational structure influences onboarding of new employees.

Interestingly, there is a modest but statistically significant trend between ensemble disagreement and ensemble prediction.
As the distribution of correlations reported in the literature is skewed in favor of positive correlations, we anticipated either no relationship between 
$\mathrm{E}[\hat{r}]$ and $\sigma(\hat{r})$, or
more uncertainty among the lower and negative correlations.
However, the positive linear trend was significant ($R\left(\mathrm{E}[\hat{r}], \sigma(\hat{r})\right) = 0.164, p < 10^{-30}$) and examining the candidate pairs in the lower quartile $Q_1$ of $\mathrm{E}[\hat{r}]$ compared with those in the upper quartile $Q_3$, we see a significant difference in disagreement (Mann-Whitney U test, $U = 531538, p < 10^{-43}$).
We illustrate these distributions in Fig.~\ref{fig:applications}B.

Our Query-by-committee style application is based upon a simple random search of the large space of untested correlate pairs.
It is unlikely that many strong correlations will be found from randomly pairing correlates.
Indeed, most predicted correlations were modest, with the distribution of $\avg{\rpre{}}$ peaked at $\approx 0$ (Fig.~\ref{fig:applications}B).
We anticipate that more effective searches of the space of untested correlations can be utilized.
That said, a number of plausible candidate correlate pairs were still found and this naive computational search is inexpensive and scalable relative to the resources required to conduct direct scientific research.

\section{Discussion}

The ability to predict correlational relationships from written descriptions has implications for the larger research problem of artificial intelligence.
Previous research has shown ways for AI systems to synthesize and predict physical relationships~\cite{bongard2006resilient,bongard2007automated,schmidt2009distilling}.
This now extends to social, economic, and other characteristic phenomena, and our work can help to further ground AI systems in the real world.

The volume of published findings continues to grow and automated tools are increasingly necessary to help scientists navigate the scientific record.
Our findings underscore the importance of data curation and standardized reporting formats.
As more published findings become computationally accessible, due to better data curation and advances in natural language processing, more of the scientific record becomes available for planning and executing research.
Research areas that follow reporting standards will benefit the most from computational tools
examining their publication record as standardization simplifies the process of extracting large training corpora.



Lastly, we caution that a predictive model such as the one proposed here can fruitfully serve as guidance to researchers conducting scientific investigations, but it is no replacement for those investigations.
Proposing and falsifying scientific hypotheses remains the gold standard of science and cannot be replaced by these models.
Instead, predictive models complement experiments and empirical findings by codifying the current state of the scientific record and providing helpful tools for researchers to handle the growth of this record.



%
%
%

\section*{Acknowledgments}

We gratefully acknowledge the metaBUS and Open Mind Common Sense projects for providing open datasets enabling this research.
This material is based upon work supported by the National Science Foundation under Grant No.\ IIS-1447634.

{\singlespacing    

\begin{thebibliography}{10}

\bibitem{jean2016combining}
N.~Jean, M.~Burke, M.~Xie, W.~M. Davis, D.~B. Lobell, and S.~Ermon, ``Combining
  satellite imagery and machine learning to predict poverty,'' {\em Science},
  vol.~353, no.~6301, pp.~790--794, 2016.

\bibitem{cooper1997evaluation}
G.~F. Cooper, C.~F. Aliferis, R.~Ambrosino, J.~Aronis, B.~G. Buchanan,
  R.~Caruana, M.~J. Fine, C.~Glymour, G.~Gordon, B.~H. Hanusa, {\em et~al.},
  ``An evaluation of machine-learning methods for predicting pneumonia
  mortality,'' {\em Artificial intelligence in medicine}, vol.~9, no.~2,
  pp.~107--138, 1997.

\bibitem{caruana2015intelligible}
R.~Caruana, Y.~Lou, J.~Gehrke, P.~Koch, M.~Sturm, and N.~Elhadad,
  ``Intelligible models for healthcare: Predicting pneumonia risk and hospital
  30-day readmission,'' in {\em Proceedings of the 21th ACM SIGKDD
  International Conference on Knowledge Discovery and Data Mining},
  pp.~1721--1730, ACM, 2015.

\bibitem{cirecsan2013mitosis}
D.~C. Cire{\c{s}}an, A.~Giusti, L.~M. Gambardella, and J.~Schmidhuber,
  ``Mitosis detection in breast cancer histology images with deep neural
  networks,'' in {\em International Conference on Medical Image Computing and
  Computer-assisted Intervention}, pp.~411--418, Springer, 2013.

\bibitem{cruz2014automatic}
A.~Cruz-Roa, A.~Basavanhally, F.~Gonz{\'a}lez, H.~Gilmore, M.~Feldman,
  S.~Ganesan, N.~Shih, J.~Tomaszewski, and A.~Madabhushi, ``Automatic detection
  of invasive ductal carcinoma in whole slide images with convolutional neural
  networks,'' in {\em Medical Imaging 2014: Digital Pathology}, vol.~9041,
  p.~904103, International Society for Optics and Photonics, 2014.

\bibitem{xu2016deep}
J.~Xu, X.~Luo, G.~Wang, H.~Gilmore, and A.~Madabhushi, ``A deep convolutional
  neural network for segmenting and classifying epithelial and stromal regions
  in histopathological images,'' {\em Neurocomputing}, vol.~191, pp.~214--223,
  2016.

\bibitem{raccuglia2016machine}
P.~Raccuglia, K.~C. Elbert, P.~D. Adler, C.~Falk, M.~B. Wenny, A.~Mollo,
  M.~Zeller, S.~A. Friedler, J.~Schrier, and A.~J. Norquist,
  ``Machine-learning-assisted materials discovery using failed experiments,''
  {\em Nature}, vol.~533, no.~7601, p.~73, 2016.

\bibitem{richards2004efficient}
G.~T. Richards, R.~C. Nichol, A.~G. Gray, R.~J. Brunner, R.~H. Lupton, D.~E.~V.
  Berk, S.~S. Chong, M.~A. Weinstein, D.~P. Schneider, S.~F. Anderson, {\em
  et~al.}, ``Efficient photometric selection of quasars from the {Sloan Digital
  Sky Survey}: 100,000 $z < 3$ quasars from {Data Release One},'' {\em The
  Astrophysical Journal Supplement Series}, vol.~155, no.~2, p.~257, 2004.

\bibitem{fiorentin2007estimation}
P.~R. Fiorentin, C.~Bailer-Jones, Y.~S. Lee, T.~C. Beers, T.~Sivarani,
  R.~Wilhelm, C.~A. Prieto, and J.~Norris, ``Estimation of stellar atmospheric
  parameters from {SDSS/SEGUE} spectra,'' {\em Astronomy \& Astrophysics},
  vol.~467, no.~3, pp.~1373--1387, 2007.

\bibitem{baldi2014searching}
P.~Baldi, P.~Sadowski, and D.~Whiteson, ``Searching for exotic particles in
  high-energy physics with deep learning,'' {\em Nature communications},
  vol.~5, p.~4308, 2014.

\bibitem{bosco2015correlational}
F.~A. Bosco, H.~Aguinis, K.~Singh, J.~G. Field, and C.~A. Pierce,
  ``Correlational effect size benchmarks,'' {\em Journal of Applied
  Psychology}, vol.~100, no.~2, p.~431, 2015.

\bibitem{bosco2015cloud}
F.~A. Bosco, P.~Steel, F.~L. Oswald, K.~Uggerslev, and J.~G. Field,
  ``Cloud-based meta-analysis to bridge science and practice: Welcome to
  metabus,'' {\em Personnel Assessment and Decisions}, vol.~1, no.~1, p.~2,
  2015.

\bibitem{collobert2008unified}
R.~Collobert and J.~Weston, ``A unified architecture for natural language
  processing: Deep neural networks with multitask learning,'' in {\em
  Proceedings of the 25th international conference on Machine learning},
  pp.~160--167, ACM, 2008.

\bibitem{mikolov2013distributed}
T.~Mikolov, I.~Sutskever, K.~Chen, G.~S. Corrado, and J.~Dean, ``Distributed
  representations of words and phrases and their compositionality,'' in {\em
  Advances in neural information processing systems}, pp.~3111--3119, 2013.

\bibitem{pennington2014glove}
J.~Pennington, R.~Socher, and C.~Manning, ``{GloVe}: Global vectors for word
  representation,'' in {\em Proceedings of the 2014 conference on empirical
  methods in natural language processing (EMNLP)}, pp.~1532--1543, 2014.

\bibitem{bengio2003neural}
Y.~Bengio, R.~Ducharme, P.~Vincent, and C.~Jauvin, ``A neural probabilistic
  language model,'' {\em Journal of machine learning research}, vol.~3,
  no.~Feb, pp.~1137--1155, 2003.

\bibitem{speer2017conceptnet}
R.~Speer, J.~Chin, and C.~Havasi, ``{ConceptNet} 5.5: An open multilingual
  graph of general knowledge,'' in {\em AAAI Conference on Artificial
  Intelligence}, pp.~4444--4451, 2017.

\bibitem{bolukbasi2016man}
T.~Bolukbasi, K.-W. Chang, J.~Y. Zou, V.~Saligrama, and A.~T. Kalai, ``Man is
  to computer programmer as woman is to homemaker? {D}ebiasing word
  embeddings,'' in {\em Advances in Neural Information Processing Systems},
  pp.~4349--4357, 2016.

\bibitem{Caliskan183}
A.~Caliskan, J.~J. Bryson, and A.~Narayanan, ``Semantics derived automatically
  from language corpora contain human-like biases,'' {\em Science}, vol.~356,
  no.~6334, pp.~183--186, 2017.

\bibitem{speer2012representing}
R.~Speer and C.~Havasi, ``Representing general relational knowledge in
  {ConceptNet} 5,'' in {\em LREC}, pp.~3679--3686, 2012.

\bibitem{seung1992query}
H.~S. Seung, M.~Opper, and H.~Sompolinsky, ``Query by committee,'' in {\em
  Proceedings of the fifth annual workshop on Computational learning theory},
  pp.~287--294, ACM, 1992.

\bibitem{krogh1995neural}
A.~Krogh and J.~Vedelsby, ``Neural network ensembles, cross validation, and
  active learning,'' in {\em Advances in neural information processing
  systems}, pp.~231--238, 1995.

\bibitem{oppenheimer2009instructional}
D.~M. Oppenheimer, T.~Meyvis, and N.~Davidenko, ``Instructional manipulation
  checks: Detecting satisficing to increase statistical power,'' {\em Journal
  of Experimental Social Psychology}, vol.~45, no.~4, pp.~867--872, 2009.

\bibitem{bongard2006resilient}
J.~Bongard, V.~Zykov, and H.~Lipson, ``Resilient machines through continuous
  self-modeling,'' {\em Science}, vol.~314, no.~5802, pp.~1118--1121, 2006.

\bibitem{bongard2007automated}
J.~Bongard and H.~Lipson, ``Automated reverse engineering of nonlinear
  dynamical systems,'' {\em Proceedings of the National Academy of Sciences},
  vol.~104, no.~24, pp.~9943--9948, 2007.

\bibitem{schmidt2009distilling}
M.~Schmidt and H.~Lipson, ``Distilling free-form natural laws from experimental
  data,'' {\em Science}, vol.~324, no.~5923, pp.~81--85, 2009.

\end{thebibliography}




\section*{Supplementary material}

# Supporting Material for "Neural language representations predict outcomes of scientific research"

James P. Bagrow[1,2,*], Daniel Berenberg[3,2], and Joshua Bongard[3,2]

[1]Department of Mathematics & Statistics, University of Vermont, Burlington, VT, United States
[2]Vermont Complex Systems Center, University of Vermont, Burlington, VT, United States
[3]Department of Computer Science, University of Vermont, Burlington, VT, United States
[*]Corresponding author. Email: james.bagrow@uvm.edu, Homepage: bagrow.com

May 17, 2018

## Contents

## List of figures

## List of tables

## S1   Datasets

We used the metaBUS data release v2.08 for our corpus of correlate pairs. These data are available for download at http://www.frankbosco.com/data/CorrelationalEffectSizeBenchmarks.html. More up-to-date data are searchable using the metaBUS web interface: http://metabus.org. We also used the 300-dimension (English) word vector representations released by the ConceptNet project called ConceptNet Numberbatch, specifically version 17.06: https://github.com/commonsense/conceptnet-numberbatch.

    The correlate texts have already been curated by the metaBUS team but we performed some further processing to connect the individual words (tokens) to terms in Numberbatch. Nonalphanumeric characters were removed, casing was removed, and text was tokenized on whitespace. Tokens were then mapped

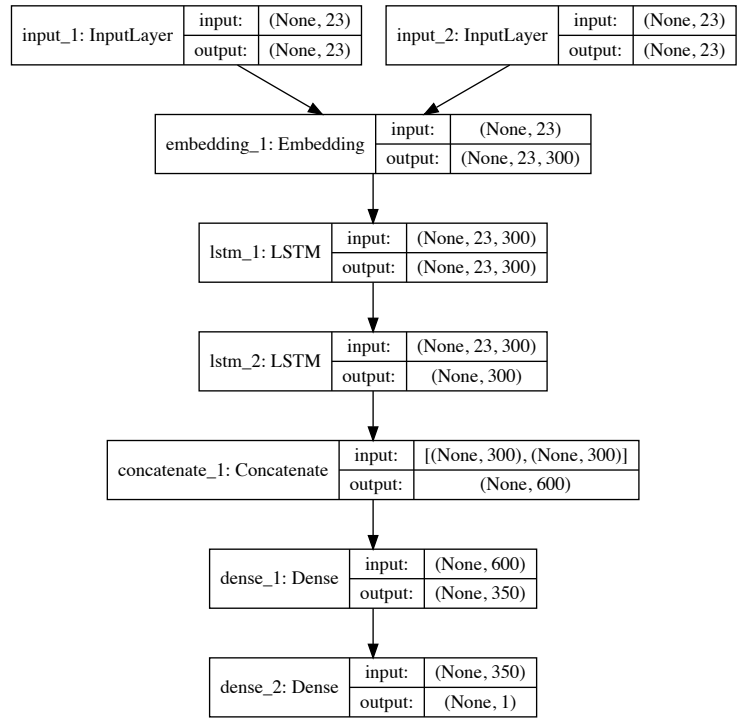

Figure S1: The neural network architecture, visualized using Keras.

to corresponding word vector indices in Numberbatch. A vector "index" of 0 was reserved for tokens in metaBUS not present in Numberbatch. The neural network is able to handle tokens outside the vocabulary of Numberbatch, although predictive performance is likely wose when many tokens are missing than when few or no tokens are missing.

## S2 Predictive models

Here we describe the neural networks used in our study.

### S2.1 Network Architecture

Our architecture takes two text sequences (the correlate pairs) and outputs their predicted correlation $\hat{r}$. The first layer consists of two inputs, one for each sequence, which lead into a static, untrainable embedding layer that translates each word in each sequence to its 300-dimensional word vector representation. Then, each word vector tensor passes through a stack of two Long Short-Term Memory (LSTM) layers of 300 units. The outputs of the second LSTM are then concatenated and sent though a dense layer of 350 units, which then feeds to a single output unit with a tanh activation. The tanh activation function constrains the network to predict $\hat{r} \in [-1, 1]$. We use ReLU activation functions In the two LSTM and single dense layers preceding the final unit.

## S2.2 Training Procedure

Models were trained for a maximum of 200 epochs with a batch size of 1024 using the Adam optimization method [1] with an MSE objective function and a learning rate of 0.001. We randomly split the metaBUS corpus into 80% training and 20% testing, and reserved 10% of the training portion as validation data. The weights of the LSTM layers were initialized using the He normal method [2]. To monitor and avoid overfitting, we apply two different methods: dropout and early stopping. At each trainable layer of the model (all excluding the embedding layer) we apply a dropout rate of 0.15, meaning at each training step approximately 15% of any given layer does not contribute to the prediction at that step, putting more pressure on each individual unit to learn valuable information [3]. Meanwhile, the early stopping mechanism monitors the loss of the model on the validation data: if the validation loss does not improve after 8 consecutive training epochs, the training is terminated and the best model so far is saved, reducing wasted computational time and helping to prevent overfitting [4]. Each of the models was trained using the Keras 2.1.6 Python library with a Tensorflow backend on an NVIDIA Tesla K80 GPU.

## S3 Mean-value baseline cannot explain neural network predictions

To determine if the neural network is simply memorizing the correlations present in the training data, we implemented a simple baseline or mean-value procedure: when predicting the correlation between pairs $c_i$ and $c_j$, simply predict the mean of all previously reported correlations involving either $c_i$ or $c_j$.

The predictive accuracy of the mean-value baseline ($R = 0.54$) is significantly lower than the neural network ($R = 0.82$) (Fig. S2). Further, the mean-value baseline is not as effective as the neural network at infilling the multi-paper correlation table shown in the main text. We compare their predictions in Figs. S3 and S4. Note that the mean-value infills are computed using the full training corpus, not the reported correlations shown in these 10 papers only. There is considerable information present in the difference between these infilled correlation tables, further illustrating the usefulness of the neural network above that of the mean-value baseline.

Of course, this mean value model can be improved by computing a weighted mean using the similarities of the correlates, for example by cosine distance between their word vector representations, but at that point it is probably more appropriate to use the neural network we applied in the main text.

## S4 Using ensemble disagreement to select candidate correlate pairs

We trained an ensemble of $N = 50$ neural networks to build the "committee" used to select candidate correlate pairs in the main text. These models were similar to the main model shown in Fig. S1 but simpler. The primary difference is that they consisted of only one LSTM layer.

To maintain a diverse ensemble of models, a common goal for ensemble learning [5, 6], we took the following steps. Each model was trained on a bootstrap replicate of the complete metaBUS dataset, meaning that an approximately $1 - 1/e$ fraction of the data of will be out-of-bag for each member of the ensemble. Next before training each model we randomized some of its hyperparameters over a range of values. Specifically we chose for each member a random number of LSTM units between 150 and 250, and a random number of dense units between 100 and 200. We also gave the LSTM layer a random dropout rate between 0.1 and 0.2, and also the dense layer's dropout rate was randomly chosen between 0.1 and 0.2. A batch size of 512 was used for training and a 10% validation sample was reserved for early stopping (3 epochs as opposed to 8 for

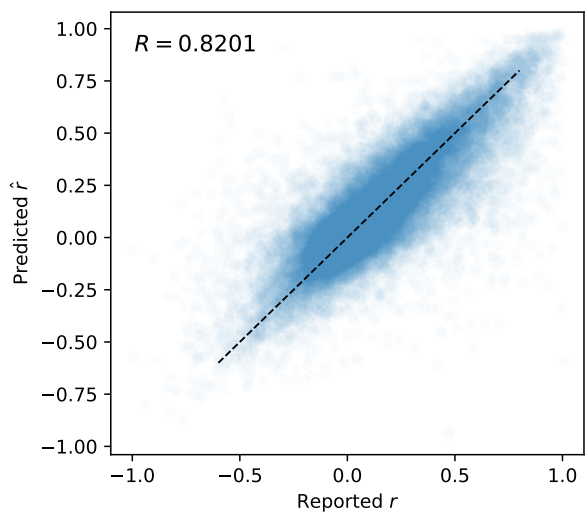
(a) Neural network.

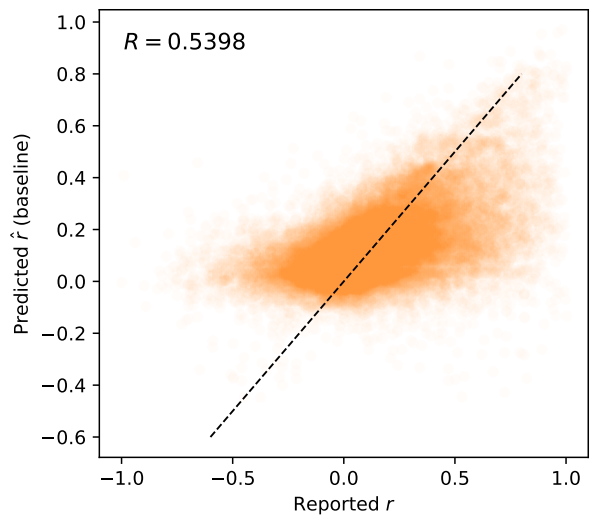
(b) Mean-value baseline

Figure S2: The mean-value baseline model (b) performs significantly worse at prediction than the neural network (a) we used in the main text. It performs especially poorly in comparison for negative correlations.

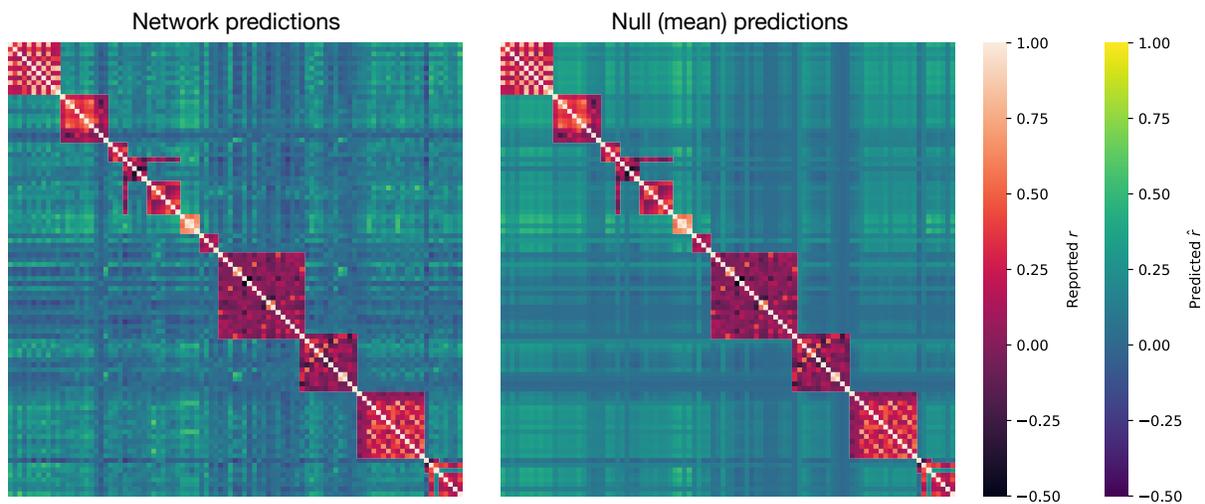

Figure S3: The mean value baseline provides less information to infill multi-paper correlation tables than the neural network.

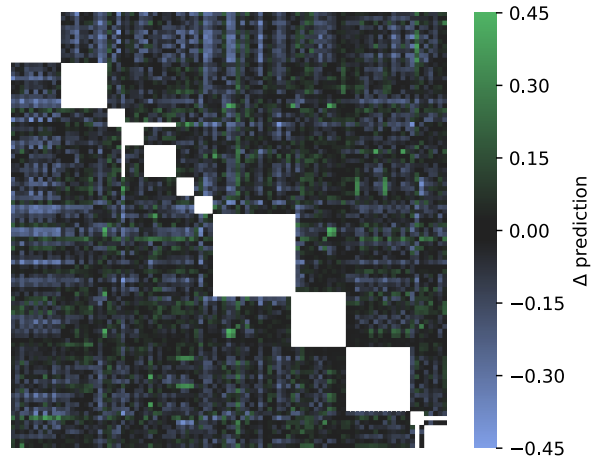

Figure S4: Difference in infilled predictions between mean-value baseline and neural network. This figure shows the difference between the matrices shown in Fig. S3. Positive values indicate the neural network predicts higher correlations than the baseline, negative values indicate the baseline predicts higher values than the neural network. We see distinct patterns in the difference, underscoring the presence of information in the neural network predictions not present in the mean-value baseline.

the main model). No activation function was used on the output unit, as opposed to the tanh function used for the main model.

Table S1 shows the correlates with the 1% most ensemble disagreement. We see a number of interesting pairs just from the small ($n = 5000$ pairs; main text) search.

Table S1: Candidate correlate pairs with ensemble disagreement in the Top 1%.

| Correlate 1 | Correlate 2 | E[$\hat{r}$] | $\sigma(\hat{r})$ |
|---|---|---|---|
| Aggression-supervisor | Video characteristics: Number of concrete statements | 0.033091 | 0.240711 |
| actor trust | Mean rating (T2) Fairness condition Middle performer | 0.324968 | 0.225283 |
| Novice completion time (Task 1, in seconds) | Organization embeddedness | -0.040126 | 0.220755 |
| Experimental 3- item loading assumption | Physical (compensation scale) | 0.167164 | 0.202105 |
| Obstructionism | Meeting format | -0.083101 | 0.199643 |
| Counterfactual thinking regarding treatment toward self | Value of surveys | -0.104113 | 0.197323 |
| Job analysis: Detail of descriptor | Mean rating (T2) Motivating condition Middle performer | 0.062453 | 0.185485 |
| Interpersonal affect | Global job satisfaction (1-5) | 0.306790 | 0.184210 |
| Counterfactual thinking regarding treatment toward self | Preview information | -0.098595 | 0.180608 |
| Anticipated regret about persistence (-10 to + 10) | Bases of power: Referent (field training) | 0.120924 | 0.180003 |
| Social forces | Desire (discretion) | 0.032081 | 0.178413 |
| Neuroticism | Instructor characteristics: Teaching skill | -0.184349 | 0.176696 |
| Behavior Commission | Instructor characteristics: Completion time on task (in minutes) | -0.095671 | 0.174231 |
| G coefficient variance component (z score) | Job Satisfaction | -0.082509 | 0.173581 |
| Behavioral intentions | Recommendations made | -0.009320 | 0.173255 |
| Manipulation check PA | Hardworking | 0.451182 | 0.172144 |
| Correlational accuracy | Satisfaction | -0.041887 | 0.171874 |
| Modem Racism Scale (MRS) | Exhaustion | -0.059882 | 0.171266 |
| Trust (T1) | Military invasive | -0.008274 | 0.170604 |
| Organizational tenure | On-time retirement | 0.105206 | 0.167623 |
| Gross domestic product | Time 2 variables: Job search behavior (unemployed group) | -0.376718 | 0.167058 |
| Mean rating (T1) Fairness condition Middle performer | Performance (Time 2) (T2) | 0.236572 | 0.166747 |
| Work satisfaction | Simple Nonstrategic Actions (P) | -0.237742 | 0.166404 |
| Competition | GRO X Gender | 0.124175 | 0.166231 |
| course-content variables: Instructor experience | Number of basic statements | -0.076577 | 0.166137 |
| Noncompliant behavior | Diary-keeping condition: Task condition | -0.167971 | 0.166024 |
| Sum of PcVc | Skill transfer | -0.039455 | 0.165213 |
| Intent to leave | Unit (Participants = fixed hourly pay system) | -0.032186 | 0.163733 |
| Recommendations made | Job analysis: Procedures for developing | 0.050139 | 0.162057 |
| Direct contact | Justification Question l | 0.140842 | 0.161316 |
| Intensity | Verdict ratings | -0.025061 | 0.161030 |

| Correlate 1 | Correlate 2 | E[$\hat{r}$] | $\sigma(\hat{r})$ |
| --- | --- | --- | --- |
| Accuracy measure: Borman's differential accuracy | Biodata inventory (Difference Score) | 0.143893 | 0.160265 |
| G coefficient variance component (z score) | Research implementation | -0.028003 | 0.159969 |
| Work intensity | Instructor characteristics: Expertise | 0.182853 | 0.159930 |
| Outcome expectancy | Emotional demands (T2) | -0.026845 | 0.159219 |
| Competency modeling: Ranking descriptor | Socialization (CPI) | 0.276968 | 0.158227 |
| Conscientiousness | Cognitive X Physical | -0.040164 | 0.157618 |
| Radiotherapy assistant | Prosocial Motivation X Manager Integrity X Dispositional Trust Propensity | -0.105287 | 0.157335 |
| Standard Occupational Classification (SOC) major group | Perceived interactional justice | -0.430987 | 0.157139 |
| Reaction to peer ratings for evaluation (wage) purposes | True/false Scoially Desirable Responding Score (Honest condition) | 0.137170 | 0.157074 |
| Average monitoring intensity | Average exam performance | -0.052568 | 0.156633 |
| Direct commitment (easy goal condition) | Skilled Trades | 0.508497 | 0.156310 |
| Organizational attract. T3 | Grip-Left (Female Sample) | 0.217855 | 0.156064 |
| Overall handshake | Feels superior | 0.344592 | 0.155691 |
| DCS: W (Time 1) | Ratings of others' contribution (RO) | -0.091074 | 0.155611 |
| Expected utility of withdrawal | Organizational Commitment Questionnaire (OCQ) Item 14 | -0.342942 | 0.154830 |
| HS/T3 | Team commitment | 0.083982 | 0.154235 |
| Performance | Supervisor: L Perf/L Inc | -0.135945 | 0.153173 |
| NEO Personality Inventory-Revised: Angry Hostility Scale (A-H) | Role perceptions Help-S (helping aimed at the supervisor) | 0.053814 | 0.153112 |
| NEO Personality Inventory-Revised: Angry Hostility Scale (A-H) | Flexibility (In-basket) | 0.147563 | 0.152931 |

# References

[1] D. P. Kingma and J. Ba, "Adam: A method for stochastic optimization," *arXiv preprint arXiv:1412.6980*, 2014. S3

[2] K. He, X. Zhang, S. Ren, and J. Sun, "Delving deep into rectifiers: Surpassing human-level performance on imagenet classification," in *Proceedings of the IEEE international conference on computer vision*, pp. 1026–1034, 2015. S3

[3] N. Srivastava, G. Hinton, A. Krizhevsky, I. Sutskever, and R. Salakhutdinov, "Dropout: A simple way to prevent neural networks from overfitting," *The Journal of Machine Learning Research*, vol. 15, no. 1, pp. 1929–1958, 2014. S3

[4] R. Caruana, S. Lawrence, and C. L. Giles, "Overfitting in neural nets: Backpropagation, conjugate gradient, and early stopping," in *Advances in neural information processing systems*, pp. 402–408, 2001. S3

[5] P. Sollich and A. Krogh, "Learning with ensembles: How overfitting can be useful," in *Advances in neural information processing systems*, pp. 190–196, 1996. S3

[6] T. G. Dietterich, "Ensemble methods in machine learning," in *International workshop on multiple classifier systems*, pp. 1–15, Springer, 2000. S3



\end{document}